\def\year{2022}\relax
\documentclass[letterpaper]{article} 
\usepackage{aaai22}  
\usepackage{times}  
\usepackage{helvet}  
\usepackage{courier}  
\usepackage[hyphens]{url}  
\usepackage{graphicx} 
\usepackage{natbib}  
\usepackage{caption} 
\DeclareCaptionStyle{ruled}{labelfont=normalfont,labelsep=colon,strut=off} 
\usepackage{soul}
\usepackage{url}
\usepackage{caption}
\usepackage{amsthm}
\usepackage{booktabs}
\usepackage{bbm}

\usepackage{setspace}
\usepackage{multirow}
\usepackage{subcaption}

\begin{document}

\title{Facilitating human-wildlife cohabitation through conflict prediction }


\author{Susobhan Ghosh$^1$, Pradeep Varakantham$^1$, Aniket Bhatkhande$^3$, Tamanna Ahmad$^3$, \\
Anish Andheria$^3$,  Wenjun Li$^1$, Aparna Taneja$^2$ , Divy Thakkar$^2$, Milind Tambe$^2$\\}
\affiliations{
$^1$ Singapore Management University\\
$^2$ Google Research India\\
$^3$ Wildlife Conservation Trust (WCT)
}

\maketitle

\begin{abstract}
    With increasing world population and expanded use of forests as cohabited regions, interactions and conflicts with wildlife are increasing, leading to large-scale loss of  lives (animal and human) and livelihoods (economic). While community knowledge is valuable, forest officials and conservation organisations can greatly benefit from predictive analysis of human-wildlife conflict, leading to targeted interventions that can potentially help save lives and livelihoods. However, the problem of prediction is a complex socio-technical problem in the context of limited data in low-resource regions. 
    
    \textbf{\em Identifying the ``right" features to make accurate predictions of conflicts at the required spatial granularity using a sparse conflict training dataset} is the key challenge that we address in this paper. Specifically, we do an illustrative case study on human-wildlife conflicts in the Bramhapuri Forest Division in Chandrapur, Maharashtra, India. Most existing work has considered human-wildlife conflicts in protected areas and to the best of our knowledge, this is the first effort at prediction of human-wildlife conflicts in unprotected areas and using those predictions for deploying interventions on the ground. 
    
\end{abstract}
\setstretch{0.96}


\section{Introduction}

India is home to some of the world's most biodiverse regions, housing numerous endemic species \cite{bharucha2002biodiversity}. \textit{Most forest areas in India are cohabited -- these are not protected areas} \cite{fsi2019}. Local communities maintain and take great care of these forests. High densities of carnivores and herbivores cohabiting with humans result in human-wildlife conflicts leading to loss of crops and cattle for humans 
, loss of wildlife, 
and in some cases, loss of human life. 
\cite{woodroffe2005people}. The number of human-animal conflicts in recent years in the state of Maharashtra, India \cite{pinjakartimes} for the years 2014-2018 ranged between 4496 and 8311 for cattle kills, 22568 and 41,737 for crop damage cases. 

\begin{figure}[ht]
    \centering
    \includegraphics[width=3.3in,height=1.2in]{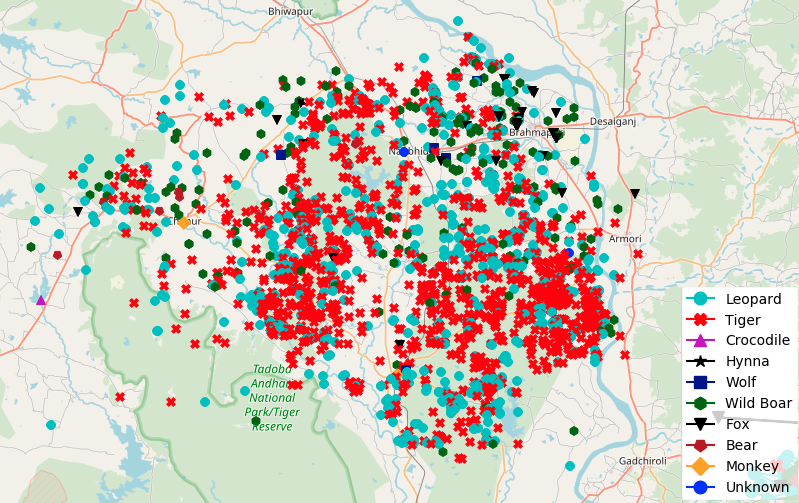}
    \caption{\small GPS plot of human-animal conflicts in the Bramhapuri Forest Division, India from 2014 to 2017}
    \label{fig:human_animal}
\vspace{-0.2in}
\end{figure}

One such region, which we focus on in this paper as an illustrative case study, is the Bramhapuri Forest Division in Chandrapur, Maharashtra, India, which is home to 2.8 tigers and 19000 humans per square kilometer. 
Studies by our on-field partner, a non-government organization (NGO), showed that more than fifty percent of the households in the Bharmapuri Forest Division had experienced crop depredation and livestock loss due to wildlife. Such conflicts  impose an economic and psychological cost on the community. Additionally, the costs also spill over to conservation efforts as in many cases these conflict situations prompt retaliatory killings of wildlife and burning of forests. Figure~\ref{fig:human_animal} shows a map of human-animal conflicts in the Bramhapuri Forest Division across 2014-17.

A big bottleneck in the mitigation of these conflicts is the lack of timely interventions. If one can predict these human-wildlife conflicts, it can help the government and NGOs execute timely interventions to reduce the loss of crops, livestock, and human-life. We aim to build AI-based solutions to help with such interventions. To that end, the main objective of this paper is to predict the intensity of human-wildlife conflicts in a particular region as an illustrative case study to learn lessons that can be utilized in other ecological domains that grapple with frequent cases of human-wildlife conflicts. In order to make such predictions, the basic requirement is the presence of conflict data over the years. Through years of interactions with the government, and conducting ground surveys, our partner NGO has collected a detailed human-animal conflict dataset since 2014.

\noindent \textbf{Contributions.} \textit{To the best of our knowledge, this is the first effort at predicting human-wildlife conflicts in unprotected areas} and this results in three main challenges. The first and foremost challenge is the need to identify the ``right'' features that will assist in the accurate prediction of conflicts. Based on observations from the data {and consultations with domain experts}, conflicts tend to happen in certain types of areas (near water bodies, low elevation areas, etc.) depending on the time period. 
Second, the conflicts are very sparse and not evenly distributed temporally and spatially. 
For instance, the dataset used in this paper has only 0.38 conflicts per month per 100 km$^2$. This poses a major challenge while trying to apply traditional machine learning tools to predict conflicts. Thirdly, for predictions to be useful, they have to be at a spatial granularity of a large village or few small villages ($\approx$ 4 km $\times$ 4 km), which is challenging. 

To address these challenges, we make the following key contributions: (i) We investigate a wide variety of features 
and conclude that simple features (like latitude, longitude, and terrain elevation) are insufficient in predicting conflicts successfully; (ii) Therefore, we move to more complex features such as satellite images and we provide a novel way of generating more training data for training Convolutional Neural Networks (CNNs) to make intensity predictions; 
(iii) To better handle sparse training data, we provide a way to apply curriculum learning and also provide a novel hierarchical classification approach.
Finally, on the real test data set, our methods provide  a prediction accuracy of 80.4\% for a spatial granularity of 4 km $\times$ 4 km. In addition to the results on a real data set, we are also in the process of deploying interventions on the ground based on our predictions (details in Section~\ref{sec:pilot}).



\section{Related Work}
While human-wildlife conflicts have been studied extensively in recent years, most of the prior work focuses on causes, mitigation and human perception of such conflicts \cite{lute2016moral, goswami2015mechanistic, treves2020myths, mccleery2009improving}. Only recently, there have been developments in successfully predicting the intensity of human-wildlife conflicts.  \cite{naha2019,sharma2020mapping} utilize land use/land cover change and map vegetation, along with other features to predict spatial patterns in human-elephant conflicts in North Bengal and human-wildlife conflicts in eastern Himalayas respectively.  However, both their areas of interest (AOI) included multiple protected areas, and their predictions highlighted the regions around these protected areas to be most prone to conflicts. In contrast, our case study does not include any protected areas in the landscape and because of this conflicts do not happen only around a few hotspots and prediction of conflicts is required at the "right" spatial granularity in the entire AOI. 

\cite{buchholtz2020using} predict wildlife-conflicts by identifying high landscape connectivity areas using circuit theory on government records and GPS tracking data from collared African elephants in Botswana. 
While their model works well with different spatial features to establish the correlation over a large time horizon, it fails to account for conflicts and their intensity across shorter time periods. In contrast, we use governmental data and publicly available satellite imagery directly to predict the conflict intensity, which is shown to be spatially robust, and also work well when temporally extrapolated. 

Raw satellite imagery has been extensively used to predict poverty \cite{jean2016combining, pandey2018multi}. \cite{bondimapping, 1562172} use segmentation techniques on satellite imagery to identify roads, forests, agriculture, etc. to further predict food market accessibility and micro-nutrient deficiency. The key distinguishing contributions of this paper are in handling the sparsity of data while predicting at the desired level of spatial granularity. 
Satellite-based remote-sensing data from public data providers like 
Google Earth Engine (GEE) \cite{gorelick2017google} has been used in several applications \cite{kumar2018google}, from crop mapping to coral reefs and landslide activity. In the domain of wildlife conservation, remote-sensing and satellite data has been used to ascertain terrain information to prevent poaching of endangered species, schedule ranger patrols \cite{fang2016deploying, xu2020stay}, and predicting poaching activities \cite{guo2020enhancing}. Unfortunately, we did not have the same data for the Bramhapuri Forest Division at the desired granularity and hence we employ satellite imagery to make conflict intensity prediction. If land use land cover data were available at the desired granularity, our contributions are directly applicable.

\section{Data collection and dataset}

\begin{figure}[ht]
    \centering
    \begin{subfigure}[b]{0.48\linewidth}
        \includegraphics[width=\linewidth]{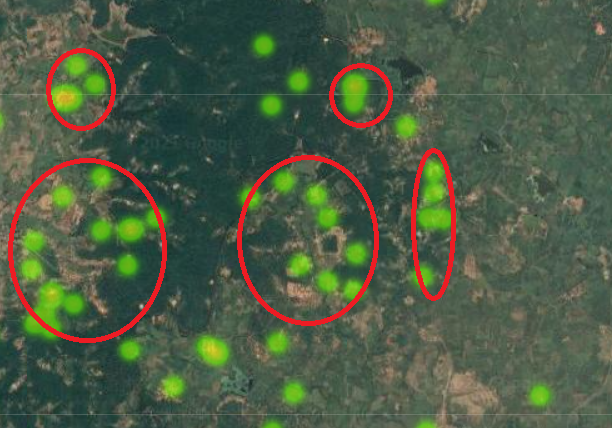}
        \caption{}
        \label{fig:red_areas}
    \end{subfigure}
    \begin{subfigure}[b]{0.48\linewidth}
        \includegraphics[width=\linewidth]{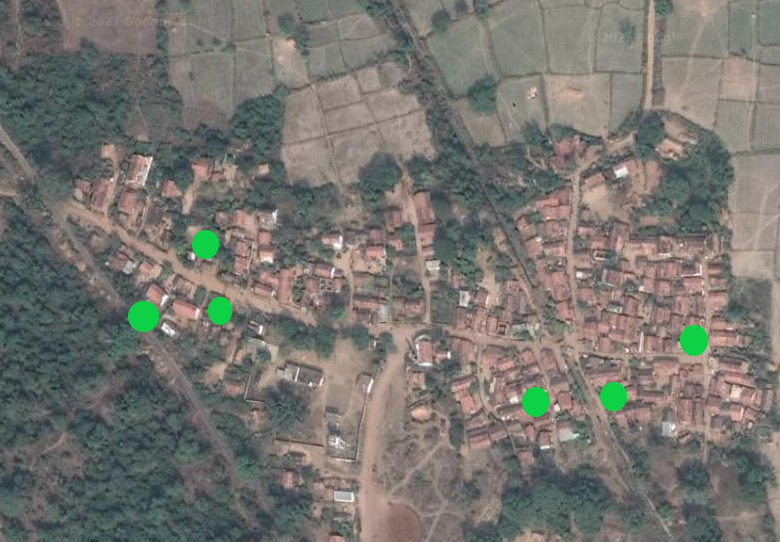}
        \caption{}
        \label{fig:settlement_c2}
    \end{subfigure}
    \caption{\small Green dots highlight individual conflicts. (a) Clustered conflicts (highlighted in red) appear near the intersection of terrains. The darker green areas denote dense forested areas, while the lighter areas have sparse vegetation. (b) Conflicts at the boundary of human settlements and forests.}
    \label{fig:conflict_terrain}
\vspace{-0.2in}
\end{figure}

One of the primary challenges with predicting human-wildlife conflicts is data collection. Since humans are severely affected during most of these conflicts, collecting data about the same becomes a sensitive issue. The government keeps records of human-wildlife conflicts through various departments. Thus accessing the data not only requires several levels of approval but also becomes a time-consuming process. However, our partnership with an NGO, accelerated the availability of data since conservation organisations maintain program records through ground surveys, and Right to Information (RTI) requests.
Our partners provided us with an anonymized conflict dataset. 
This dataset comprised of 2628 cattle-animal and human-animal conflict records from 2014 to 2017, each detailing the cattle or human being killed or injured, the attacking animal (tiger, boar, etc.), the approximate GPS coordinates of the conflict occurrence, the date of the conflict, the village details (range, round, beat, village name, and compartment number). Three animals, namely tigers, leopards, and wild boars, were involved in a majority of the human-animal conflicts, as visualized in Figure~\ref{fig:human_animal}. Since this is a real-world dataset, we had to clean and pre-process the dataset.  After the data cleaning process, we ended up with 2196 conflict records, each detailing the GPS coordinates of the conflict and date of the conflict, alongside information about the animal involved, the cattle or human being killed or injured, and the village where the conflict occurred.


\section{Problem Definition}
We have filtered past data of conflicts, ${\cal C}$ where each conflict $i \in {\cal C}$ occurs at a location, $l_i$ (latitude, longitude) and during a time period, $t_i$ (e.g., Feb 2015). Using the past conflict data for training, the objective is to predict conflict intensity, $y_r$ (e.g., low, medium, high), in a given region $r$ (defined as a continuous area of land covering a large village or few small villages) of size roughly $m$ km $\times$ $n$ km, \textit{during a different time period than the one for which conflict data is provided}.

In order to predict conflict intensity, we first convert the given conflict data into a learning problem with $R$ training examples $\{(<x_r,t_r>,y_r)_{r \in R}\}$, where:\\
\noindent \textbf{--} $x_r$ represents the features corresponding to a region $r$ of size $m$ km $\times$ $n$ km.\\
\noindent \textbf{--} $t_r$ is time period of interest (e.g., February of 2015) in the training data.\\
\noindent \textbf{--} $y_r$ is the intensity of conflicts given by $f(\sum_{i \in {\cal C}} I_{l_i, r})$, where $f(.)$ maps the number of conflicts to an intensity (e.g., low, medium, high in case of classification and the actual number of conflicts in case of regression) and $I$ is the indicator function that is 1 if $l_i \in r$ and 0 otherwise. 

Formally, the objective is to build a predictor, $\mathcal{P}$ so as to minimize the loss between the predicted intensity, $\mathcal{P}(x_r,t_r)$ and ground truth intensity, $y_r$ for a test data set (where $t_r$ will be for the time period of test dataset).  There are three key challenges in building such a predictor:\\ 
\noindent \textbf{\em Challenge 1}: Identifying the features to be considered for each region, i.e., $x_r$ to accurately predict $y_r$.\\ 
\noindent \textbf{\em Challenge 2}:  Predicting accurately with a few training examples and a significant class imbalance. \\
\noindent \textbf{\em Challenge 3}: Predicting conflict intensities for regions at the right spatial granularity ($\approx$ 4 km $\times$ 4 km).

\section{Prediction of Conflict Intensity}

Towards addressing the three challenges, we first investigate different types of features and identify the ones that provide the highest accuracy. We then provide our key technical contributions that handle the sparsity of training examples and provide predictions at the desired spatial granularity of regions.    
To address \textbf{\em challenge 1}, we work through a progression of features from simple to complex. We begin with a fixed set of regions (obtained through clustering of conflicts) and use the region identifier (cluster number) as $x_r$. This is described in the first part of Section~\ref{sec:clustering}. However, since having just a region identifier does not capture the connectivity between regions, we compute an embedding for a region determined based on the connections to other nearby regions in the later part of Section~\ref{sec:clustering}. Then, to evaluate the importance of terrain properties, we use the elevation of the region (only data available at the right level of granularity) in Section~\ref{sec:spatial}. Finally, in Section~\ref{sec:satimg}, we employ satellite imagery for a given region to not only capture context (e.g., forests, water sources, croplands, settlements, roads, etc.), but also the neighborhood of the region. The features that are common for all sections (\emph{train} and \emph{test} datasets) are mentioned below and the specialized features for each prediction method are mentioned in the corresponding sections. Depending on whether we are training or testing, $t_r$ is different. $t_r$ is the month and year for a specific example in training/testing data. 
\begin{table}[!h]
  \centering
  \vspace{-0.15in}
  {\small \begin{tabular}{c|c}
    \emph{train} & 2014-16 \\
    \hline
    \emph{test} & 2017
  \end{tabular}}
  \vspace{-0.15in}
\end{table}
\subsection{Implementation Details}
The models used in this paper were trained and evaluated on a machine with an Intel Xeon E5-2630 v4 processor, 256 GB RAM, and 8 RTX 2080 Ti GPUs, running Ubuntu 18.04.2 LTS. All the experiments were run on python 3.8. We used \emph{scikit-learn} for regression tasks, while we used \emph{pytorch} for training CNN-based methods. All the regression and simple classification methods were trained and evaluated in under a few minutes. All the CNN-based methods were trained and evaluated in under 48 hours.

\noindent The models and their implementations used in this paper can be found here \footnote{https://www.dropbox.com/s/46tpp6bzosisxvy/human-wildlife-conflict-code.zip?dl=0}. We do not provide any conflict data, or any trained model, as the real data is classified, and the trained models would easily give away the distribution of endangered animals involved in conflicts (since the AOI is mentioned in the paper). However, we do add dummy csv files with some dummy conflict data, which can be filled with the real data to replicate the results. \emph{The specific hyper-parameters used by each prediction method are mentioned in their respective sections}.

\subsection{Prediction with region identifiers}
\label{sec:clustering}
\vspace{-0.1in}
\begin{table}[!h]
  \centering
  {\small \begin{tabular}{c|c}
    $m, n$ & Mean $\approx$ 18 $\times$ 18\\
    \hline 
    $x_r$ & Cluster ($lat, long, elev$) region identifier or embedding\\
    \hline
    $y_r$ & Frequency of conflicts
  \end{tabular}}
  \vspace{-0.1in}
\end{table}


First, we employ region identifiers as $x_r$. The actual regions can either be obtained by equally dividing the overall area into regions or by clustering conflicts. In this section, we describe and provide regression results ($f(.)$ maps to the actual count of incidents) for the latter case as it performed better than equal size regions.

We applied K-Means on the GPS coordinates 
of the conflicts and generated an elbow curve. 
After looking at the elbow curve and analyzing the performance, we 
and set the number of clusters as $k=38$ and labeled all the data points. 
Using this labeled data, we group conflicts for each cluster, for each month. We generated a dataset using this cluster information, which we refer to as NWA. We then applied different regression methods, namely linear regression \cite{kenney1962linear}, ridge regression \cite{hoerl1970ridge}, stochastic gradient descent (SGD) based linear regression \cite{zhang2004solving}, and Multi-Layer Perceptron based regression to solve our prediction problem (using default hyper-parameters provided in \emph{scikit-learn}). 
We used the number of conflicts as our target variable $y_r$, the month and year as the temporal variables $t_r$, and the region identifier (cluster number obtained using $lat$, $long$ of conflicts) $x_r$. Since our models are going to be used to predict for the future, throughout the paper, we evaluate our methods by extrapolating the data. \emph{To that end, we train on the conflict data from 2014-16, and test on the conflict data from 2017}. Table~\ref{tab:regression_acc} (\emph{Standard} column) compiles the best ${R}^2$ scores for each regression model on the test set.
\begin{table}[ht]
\centering
{\small \begin{tabular}{|p{1.5cm}||c|c|p{2cm}|} 
 \hline
  Regression Type & Standard & Node2Vec & Node2Vec-Elevation)\\
  \hline
  \hline
  Linear & -0.050 & 0.050 & 0.082 \\
  Ridge & -0.050 & 0.025 & 0.082\\
  MLP & -0.045 & -0.017 & -0.355\\
  SGD & -0.053 & 0.000 & -0.021\\
 \hline
 \end{tabular}}
\vspace{-0.1in}
 \caption{\small $R^2$ scores for regression methods}
  \label{tab:regression_acc}
\vspace{-0.1in}
\end{table}

\begin{table}[ht]
    \centering
    {\small \begin{tabular}[width=\linewidth]{|c||c|c||c|c|}
    \hline
    Model & \multicolumn{2}{c||}{Node2Vec} & \multicolumn{2}{p{2.1cm}|}{Node2Vec-Elevation}\\
    \cline{2-5}
    & 5 & 3 & 5 & 3\\
    \hline
    \hline
    Logistic Regression & 48.2 & 57.0 & 60.9 & 59.0 \\
    \hline
    MLP & 46.4 & 56.1 & 59.5 & 58.9 \\
    \hline
    SVM & 50.8 & 60.5 & 60.9 & 61.9 \\
    \hline
    \end{tabular}}
\vspace{-0.1in}
        \caption{\small Accuracy (\%) for classification models}
    \label{tab:cluster_classifiers}
\vspace{-0.15in}
\end{table}
The low $R^2$ scores indicate our zone representation did not capture the adjacency information, which would essentially capture the possibility of animals migrating to nearby areas. We therefore employed node2vec \cite{grover2016node2vec} 
and generate embeddings (of size 128). We retrained all the regression models using this new representation, where $x_r$ is the embedding of the region. Table~\ref{tab:regression_acc} (\emph{Node2Vec} column) summarizes the $R^2$ scores of the models with node2vec representation, which shows improvement in performance with scope for improvement.

We also tried including the wild animal involved as a feature to predict  conflicts. 
However, this results in 90\% of regions having a total of zero conflicts, which biases the regression models to predict zero most of the time.
In contrast, our initial NWA dataset had 58\% of the dataset full of zeros. Citing this imbalance, we decided against using wild animal information in our future experiments.



\subsection{Prediction with Terrain Features}
\label{sec:spatial}
\begin{table}[!h]
  \vspace{-0.1in}
  \centering
  {\small \begin{tabular}{c|c}
    $m, n$ & Mean $\approx$ 18 $\times$ 18\\
    \hline 
    $x_r$ & Cluster ($lat, long, elev$) region identifier or embedding\\
    \hline
    $y_r$ & Frequency / Intensity of conflicts (5 and 3 classes)
  \end{tabular}}
  \vspace{-0.1in}
\end{table}
We also explored terrain features, which play an important role in the location and movement of animals. Since land-use, land cover data was unavailable at the required granularity, 
we incorporated elevation data and ran the clustering technique with GPS coordinates and elevation data, and re-labelled the datasets. 
Table~\ref{tab:regression_acc} shows that including elevation data improves $R^2$ scores in most cases.

However, due to the poor performance of regression methods and the feedback from our on-field experts that a class-based prediction for conflict areas (like low, medium, or high conflict areas) was sufficient (predicting the actual number of conflicts was not as crucial), we move away from regression-based methods and explore classification methods. We bucketed the total conflicts for each record in our NWA dataset into five ([0], [1-3], [4-6], [6-9], [10+]) and three classes ([0], [1-9], [10+]). We then applied logistic regression \cite{hosmer2013applied}, support vector machine (SVM) \cite{suykens1999least}, and multi-layer perceptron (MLP) \cite{glorot2010understanding} based classifiers on this dataset. 
Table~\ref{tab:cluster_classifiers} summarizes the accuracy of the three classifiers. Highest accuracy of 60.5\% is observed when using node2vec without elevation data, on the dataset having three classes. Including elevation data during clustering in the classification task did positively impact the accuracy of the models, without significant boost in performance of the models. 

In conclusion, our first contribution goes to highlight that simple features (like latitude, longitude, and elevation) are not sufficient to predict conflicts with a high degree of accuracy, either using classification or regression. This coupled with certain patterns in conflicts (as observed in Figure~\ref{fig:conflict_terrain}) motivates us to employ satellite imagery to capture terrain and adjacency information to improve prediction accuracy. 

\subsection{Prediction with Satellite Imagery}
\label{sec:satimg}
As indicated, our focus is on identifying conflicts in unprotected areas and this can result in conflicts at many different places including but not limited to intersections of different terrains, lower elevation regions, water bodies, boundaries of villages and forests, etc. 
We incorporate this by using available true-color satellite imagery. We use ESRI's Satellite Imagery API \cite{Satellit59:online} and Google Static Maps API \cite{Overview61:online} to get the base maps for our AOI. This allows  predicting conflicts for regions of our choice, instead of only static cluster regions generated by clustering methods with previous methods. 

Given their effectiveness in dealing with image data, Convolutional Neural Networks (CNNs)~\cite{lecun2015deep} trained on sufficient size grids (to capture the context of the region) of the satellite imagery would present themselves as a natural option. Unfortunately, given a total AOI of approximately 132 km $\times$ 121 km, either the number of training examples is small or the size of the region is too small for CNNs to identify the context (forests, water sources, croplands, settlements, roads, etc.).

To address \textbf{\em challenge 2}, a key insight (and our second contribution) that we employ is to divide the satellite imagery of AOI into \textbf{overlapping} equal-sized grids. By controlling the overlap (explained in detail below), it is feasible to generate more images to train the CNN. 

\noindent \subsubsection{Generating Dataset}
\label{sec:gen_dataset}
There are two major issues related to \textbf{\em challenge 2}. First, if the satellite image is just partitioned into square regions once, it would result in a few images - not enough to train a CNN-based network. Second, since most of these regions are likely to have zero conflicts, they would skew the dataset, as seen in Section~\ref{sec:clustering}. 

We address the first issue by taking overlaps on the satellite imagery i.e. by shifting the square regions by some offset after one round of partitions. This potentially allows the creation of an unlimited number of images (or data points). However, we limit it to five offsets along the longitude and latitude, creating  five times the number of images from one pass of partitioning over the full satellite image. 
The dataset thus created, referred to as SAT0$(k)$, has records that contain the satellite image of an AOI in a $k$ km $\times$ $k$ km radius ($x_r$), the temporal features - month and year ($t_r$), and the corresponding number of conflicts in that AOI during that month and year ($y_r$). To address the second issue, we create new versions of our dataset SAT0($k$) by removing all areas where conflicts haven't been reported in the entire three-year time frame (2014-17). We label this dataset as SAT1($k$) and use it in our experiments.

\begin{figure}[!t]
    \centering\
    \begin{subfigure}[b]{0.155\linewidth}
        \includegraphics[width=\linewidth]{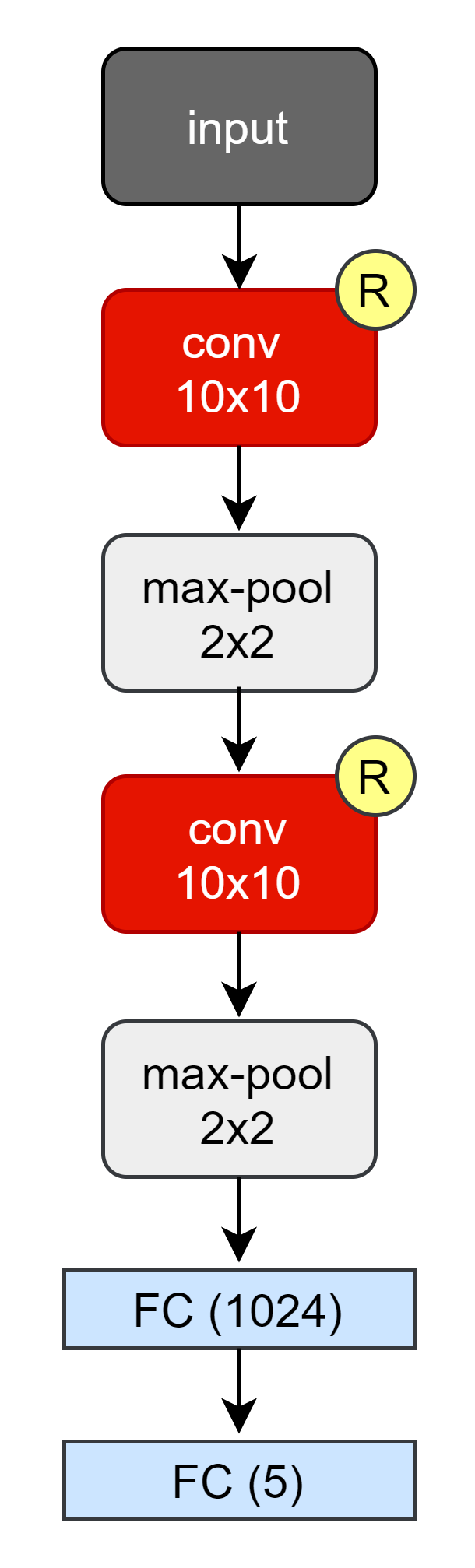}
        \caption{N1}
        \label{fig:N1}
    \end{subfigure}
    \begin{subfigure}[b]{0.155\linewidth}
        \includegraphics[width=\linewidth]{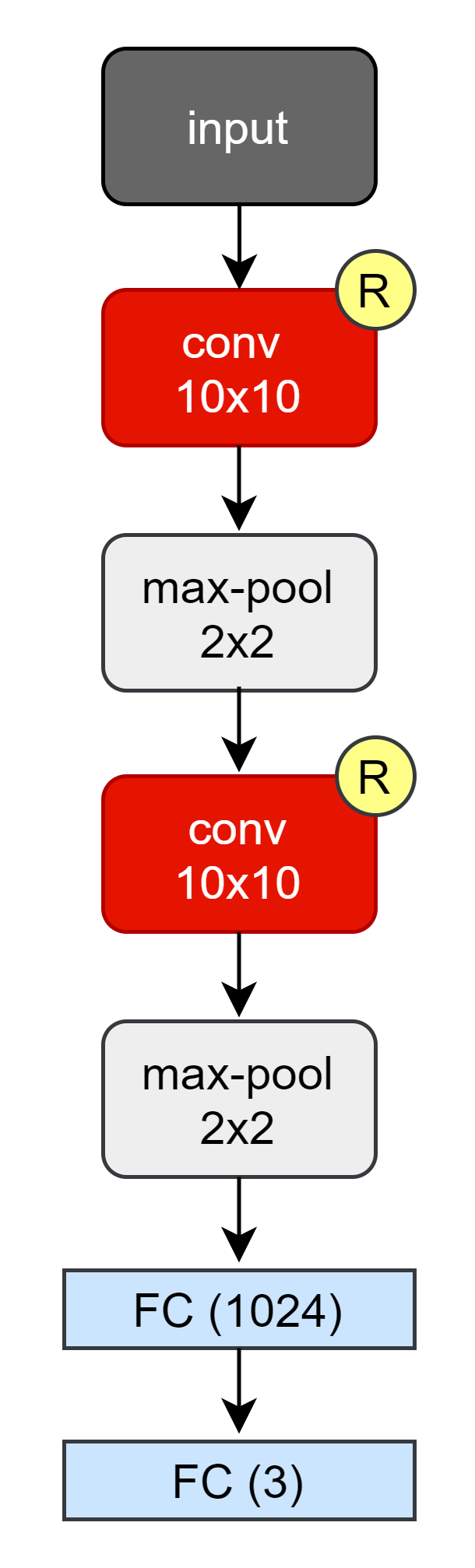}
        \caption{N2}
        \label{fig:N2}
    \end{subfigure}
    \begin{subfigure}[b]{0.155\linewidth}
        \includegraphics[width=\linewidth]{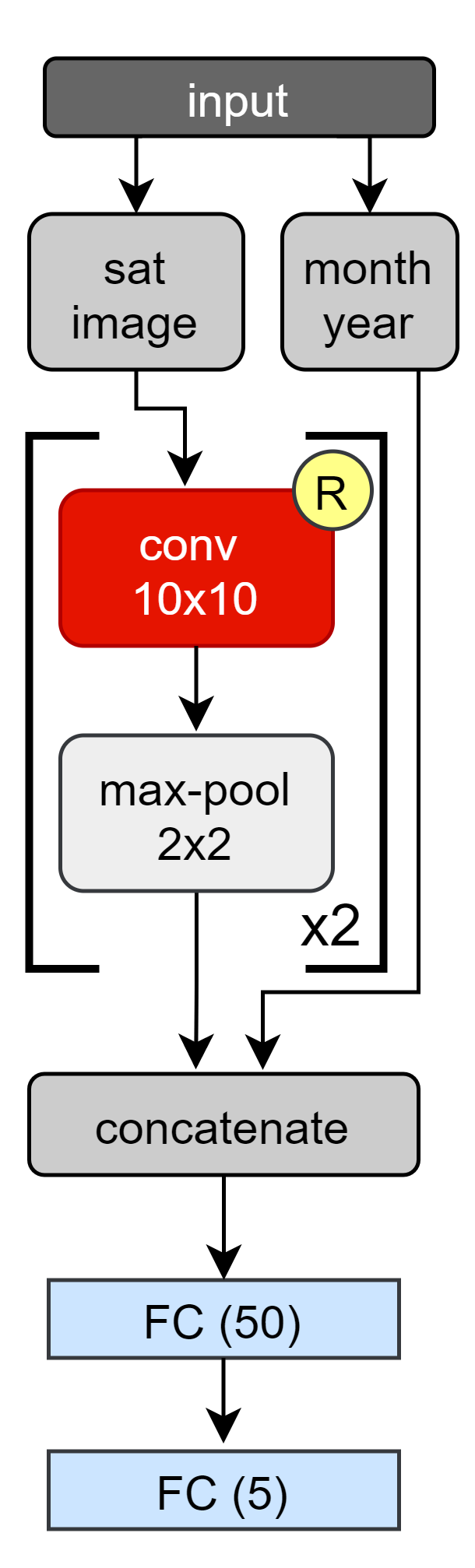}
        \caption{N3}
        \label{fig:N3}
    \end{subfigure}
    \begin{subfigure}[b]{0.155\linewidth}
        \includegraphics[width=\linewidth]{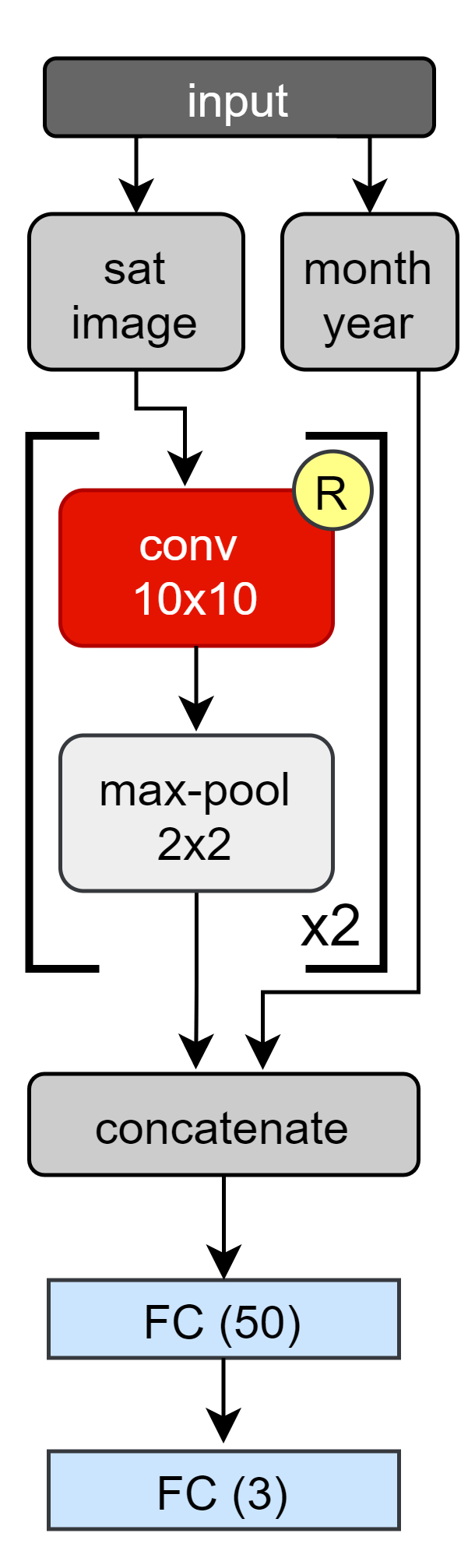}
        \caption{N4}
        \label{fig:N4}
    \end{subfigure}
    \begin{subfigure}[b]{0.155\linewidth}
        \includegraphics[width=\linewidth]{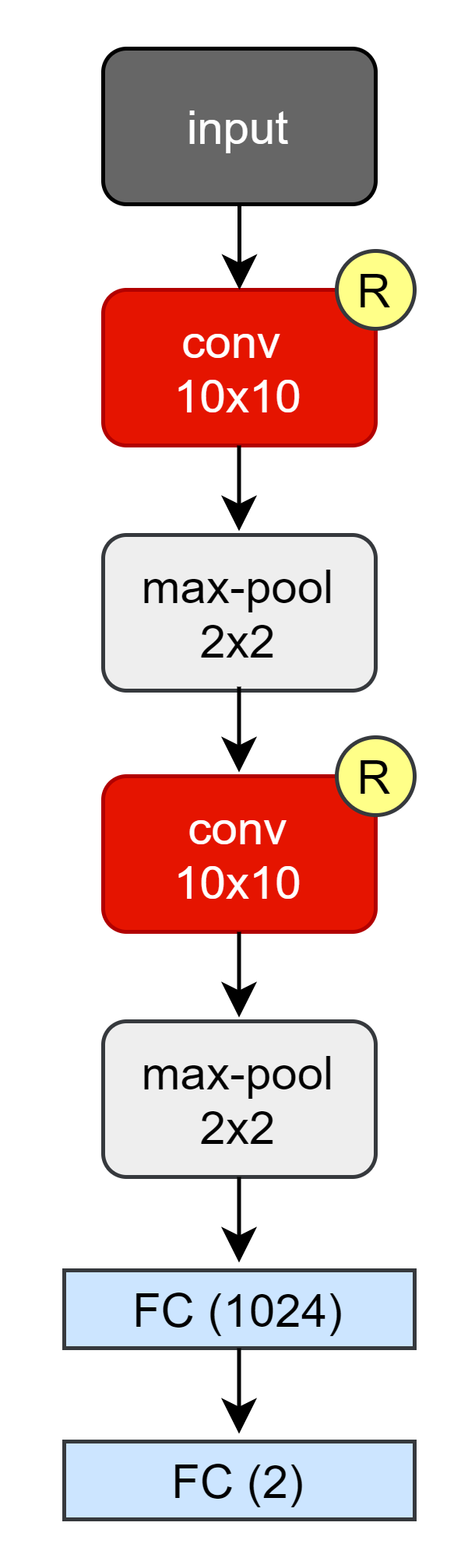}
        \caption{N5}
        \label{fig:N5}
    \end{subfigure}
    \begin{subfigure}[b]{0.155\linewidth}
        \includegraphics[width=\linewidth]{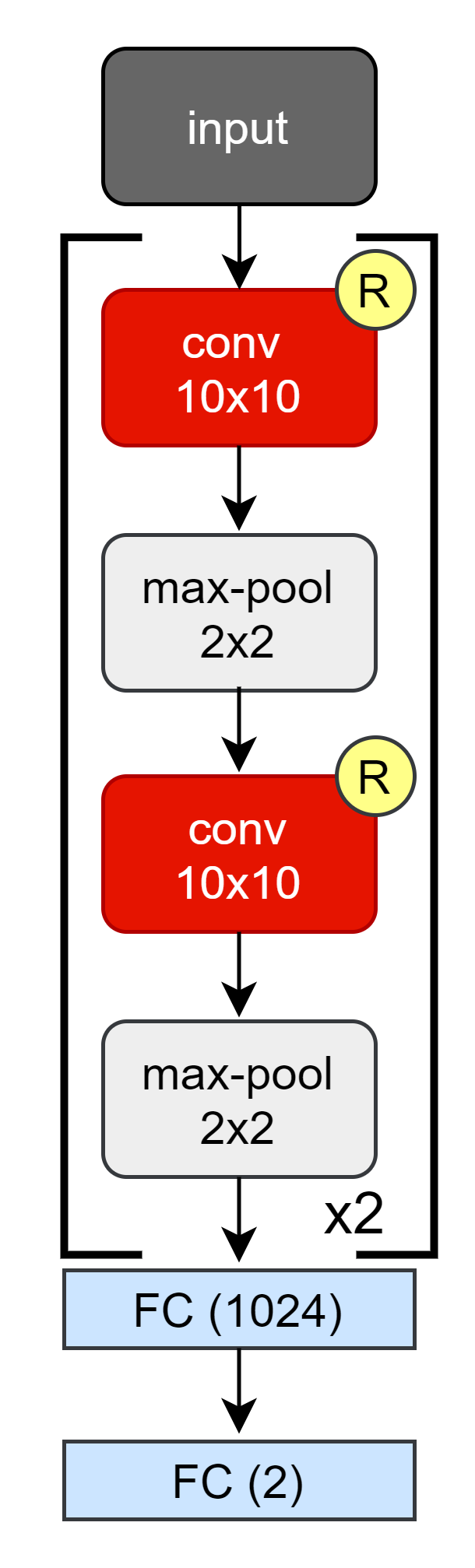}
        \caption{N6}
        \label{fig:N6}
    \end{subfigure}    
    \caption{\small CNN-based network architectures used in Section \ref{sec:cnn_models} [(a)-(d)], Section \ref{sec:curriculum} [(e)-(f)] and Section \ref{sec:hierarchical} [(e)-(f)]. The notation: conv $(d \times d)$ refers to a convolutional layer with $d \times d$ filters; max-pool ($s \times s$) denotes max-pooling operation with stride $s$ - a stride of 2 means the inputs are downsampled by a factor of 2; FC($n$) denotes a fully connected layer with $n$ output features; R represents the relu activation function.}
    \label{fig:CNNs}
  \vspace{-0.25in}
\end{figure}

\noindent \textbf{Solutions Overview: } Now that we have a dataset with sufficient satellite images, we provide a series of approaches to address \textbf{\em challenge 2} and \textbf{\em challenge 3}. First, we provide a CNN based learning model (for SAT1(10)) that is able to achieve high accuracy, precision and recall values. However, it achieves less than 60\% accuracy on SAT1(4), which is the granularity of interest. To address this, we propose a curriculum learning method that reduces the granularity gradually (SAT1(10)$\rightarrow$SAT1(8)$\rightarrow$SAT1(4)) while training, rather than directly going to SAT1(4). Such an approach is shown to learn better due to understanding the contour of the solution space in easier problems first. This improves the accuracy substantially, however, the precision and recall values are very low for SAT1(4) due to the data sparsity for non-zero conflicts. To address this, we propose a hierarchical classification approach that not only employs the decrease in granularity while training, but also explicitly emphasizes training on non-zero conflict areas. 
This ensures better overall training and as shown in our results (Table~\ref{tab:full_acc}) achieves high accuracy, precision and recall values.   
\subsubsection{CNN-based Learning Model}
\label{sec:cnn_models}
We use the datasets generated above to predict the nature of conflicts in a particular AOI. To do so, we first bucket and label the conflicts into multiple classes (we use two variants - five classes and three classes, similar to the ones described in Section~\ref{sec:clustering}?). Then we take the images as grayscale input (1 channel) and pass the temporal features $t_r$ (month and year) after normalization in the other two channels - making it a 3-channel image. The images are then passed to train CNN-based prediction networks. The corresponding networks labeled N1 (for five classes) and N2 (for three classes) are described in Figure~\ref{fig:N1} and~\ref{fig:N2}.
 \begin{table}[!h]
   \vspace{-0.1in}
  \centering
  {\small \begin{tabular}{c|c}
    $m, n$ & 10 $\times$ 10\\
    \hline 
    $x_r$ & Satellite imagery\\
    \hline
    $y_r$ & Intensity of conflicts (5 and 3 class)
  \end{tabular}}
  \vspace{-0.13in}
\end{table}
We also train two multi-headed networks labeled N3 (for five classes) and N4 (for three classes), described in Figure~\ref{fig:N3} and~\ref{fig:N4}. These take only the original grayscale 1-channel images as input to the convolution layers, and later concatenate the temporal features of month and year with the output of convolution layers, before passing it to the fully connected layers. For training all our networks, we use the Adam optimizer \cite{kingma2014adam}, with a learning rate of 0.0001. We use weighted cross-entropy loss, and we train for 300 epochs. The performance of our CNN-based models on these datasets are summarized in Table~\ref{tab:classifiers}. We find that our N2 model reaches an accuracy of 82.2\% on the SAT1(10) dataset. We highlight this model, as it has high precision and recall values. We also observe that networks trained for five-class classification tasks perform poorly; hence we do not show five-class classification results in our future sections.

\begin{table}[ht]
    \centering
    {\small \begin{tabular}[width=\linewidth]{|c||c|c|c|c|}
    \hline
    Model (Classes) & N1 (5) & N2 (3) & N3 (5) & N4 (3)\\
    \hline
    \hline
    Accuracy & 66.3 \% & \textbf{82.2} \% & 62.8 \% & 73.1 \%\\
    \hline
    Precision & 0.44 & \textbf{0.73} & 0.38 & 0.58\\
    \hline
    Recall & 0.38 & \textbf{0.81} & 0.44 & 0.69\\
    \hline
    \end{tabular}}
    \vspace{-0.1in}
    \caption{\small Results for non-zero conflict classes on SAT1(10) }
    \label{tab:classifiers}
    \vspace{-0.2in}
\end{table}

\begin{table}[ht]
    \centering
    {\small \begin{tabular}[width=\linewidth]{|c||c|c|c|}
    \hline
    Offset & O1 & O2 & O3\\
    \hline
    \hline
    Accuracy & 73.7\% & 74.0\% & 80.0\%\\
    \hline
    Precision & 0.63 & 0.65 & 0.71\\
    \hline
    Recall & 0.71 & 0.71 & 0.80\\
    \hline
    \end{tabular}}
    \vspace{-0.05in}
        \caption{\small Results on SAT1(10) with  offsets (non-zero classes)}
            \label{tab:offset_acc}
        \vspace{-0.1in}
\end{table}


In order to test the robustness of our trained N2 based model with respect to a spatial shift, we generated a new set of data containing images with offsets that were not used during training, namely O1, O2 and O3. O1 is a testing dataset generated with an offset of 1.11 km, while O2 and O3 are generated with offsets of 2.77 km. They utilize all the conflict data from 2014 to 2017, except for O3, where we take only the conflict data from 2017 to generate the data points. We report our test results in Table~\ref{tab:offset_acc}. The good performance of our model showcases that our trained model is robust to spatial change in input satellite imagery.

Unfortunately, even with the N2 model, we do not cross 60\% accuracy on the SAT1(4) dataset (\textbf{\em challenge 3}). This is mainly due to the sparsity of medium or high conflict intensities in the dataset. Towards addressing this drawback, we introduce our third set of contributions on curriculum learning and an extension of curriculum learning to deal with sparse data and class imbalance. 

\subsubsection{Curriculum Learning}
\label{sec:curriculum}
Curriculum Learning (CL)~\cite{bengio2009curriculum} employs a curriculum where predictors are trained initially on easy examples and then moved to difficult examples. As highlighted in the paper~\cite{bengio2009curriculum}, a well-chosen curriculum can serve as a continuation method~\cite{allgower2012numerical}, i.e., can help to find a better local minima of a non-convex training criterion. It has been shown to improve not only training accuracy but also generalization ability.  In this paper, we train the neural network by employing a curriculum of region sizes.
 \begin{table}[!h]
  \centering
  \vspace{-0.05in}
  {\small\begin{tabular}{c|c}
    $m, n$ & 10 $\times$ 10, 8 $\times$ 8, and 4 $\times$ 4\\
    \hline 
    $x_r$ & Satellite imagery\\
    \hline
    $y_r$ & Intensity of conflicts (3 and 2 class)
  \end{tabular}}
  \vspace{-0.13in}
\end{table}

We first create three subsets of regions in a sequence,  with each subsequently numbered subset having images that cover a smaller area as compared to the previous. We hypothesize that it is easier to predict conflicts in a much larger area (i.e. at a macro-level) than in a smaller area (i.e. at a micro level), due to the sparsity of non-zero conflict regions. With that belief, we start training with the images covering larger areas of 10 km $\times$ 10 km (SAT1(10)), and later include images which cover a 8 km $\times$ 8 km area (SAT1(8)) and finally we consider images which cover a 4 km $\times$ 4 km area (SAT1(4)). During the training process, we add the subsequent subsets (with smaller regions) to our training dataset when the validation accuracy of the model on the previous set does not continuously increase for five epochs (determined empirically). We evaluate the performance of the curriculum learning model for both three-class and two-class classification problems. To that end, we utilize three networks, namely N2 (3 class), N5 and N6 (2 class), as detailed in Figure \ref{fig:N2}, \ref{fig:N5}, and \ref{fig:N6} respectively.
\begin{table}[ht]
    \centering
    {\small \begin{tabular}[width=\linewidth]{|c||c|c|c|c|}
    \hline
    Model (Classes) & N2 (3) & N5 (2) & N6 (2)\\
    \hline
    \hline
    4$\times$4 Accuracy & 77.3\% & 77.9\% & 82.2\%\\
    \hline
    4$\times$4 Precision & 0.24 & 0.26 & 0.32\\
    \hline
    4$\times$4 Recall & 0.25 & 0.27 & 0.17\\
    \hline
    \hline
    FCL Accuracy & 75.1 \% & 76.4\% & 77.2\%\\
    \hline
    FCL Precision & 0.49 & 0.54 & 0.60\\
    \hline
    FCL Recall & 0.46 & 0.57 & 0.35\\
    \hline
    \end{tabular}}
    \vspace{-0.05in}
    \caption{\small Results at the 4 km $\times$ 4 km level with CL and FCL.}
    \label{tab:curriculum}
    \vspace{-0.1in}
\end{table}
Table~\ref{tab:curriculum} summarizes the performance of the models. Here are the key observations:\\
\noindent \textbf{--} Across all the models, the best accuracy obtained with a CNN model on 4 km $\times$ 4 km was less than 60\%. However, with CL, this improves to at least 77.3\% for N2, N5 and N6 models for SAT1(4).\\
\noindent \textbf{--} The full system results (averaged over all granularities), referred to as FCL are also quite high with respect to the accuracy, and moderate to low with respect to the precision and recall.\\
\noindent \textbf{--} Unfortunately, the higher accuracy values are obtained at the cost of lower precision and recall values (maximum of 0.32) for SAT1(4).\\
The poor precision and recall values are due to a significant class imbalance (significantly many zero-conflict regions $\approx$90\%-95\% and few non-zero conflict regions) in SAT1(4). Next, we provide a hierarchical classification approach that builds on the idea of gradual decrease in granularity. 

\subsubsection{Hierarchical Classification}
\label{sec:hierarchical}


To address the class imbalance issue, the first key insight we employ is to train separately on the sparse non-zero conflict regions, so that the predictor weights do not get overwritten when learning from the many zero-conflict regions. To explicitly focus on the non-zero conflict regions, we propose a step-wise hierarchical prediction framework, which uses a combination of CNN-based predictors trained at different granularities to predict conflicts.   Specifically, in this case, we first train on a higher granularity dataset, SAT1(10) or macro level, and focus the training for SAT1(4) or micro level on those 10 km $\times$ 10 km regions with high conflict intensity. 
It is important to note that only a subset of the SAT1(4) dataset is used to train the micro-level network due to this hierarchy.

\begin{table}[!h]
  \centering
  \vspace{-0.1in}
  {\small \begin{tabular}{c|c}
    $m, n$ & 10 $\times$ 10, and 4 $\times$ 4\\
    \hline 
    $x_r$ & Satellite imagery\\
    \hline
    $y_r$ & Intensity of conflicts (3 and 2 class)
  \end{tabular}}
  \vspace{-0.1in}
\end{table}

This hierarchy enables us to make use of our best performing N2 network at the macro-level. For the micro level, we train a host of CNN-based prediction networks, namely N2, N5 and N6 (detailed in Figure~\ref{fig:N2}, ~\ref{fig:N5}, and \ref{fig:N6} respectively), and AlexNet \cite{krizhevsky2017imagenet}. The dataset is only generated using satellite imagery, and the images are clipped at a resolution of 224 $\times$ 224 pixels. 

We test our hierarchical classification model on the conflict data from 2017, and the results obtained are summarized in Table~\ref{tab:full_acc}. Here are the key observations:\\
\noindent \textbf{--} N6 model provides the best results for both 4x4 and full system accuracy, precision, and recall. The results are substantially better than those obtained with CL.\\ 
\noindent \textbf{--} The best model for 3 classes, N2  has low precision and recall values for SAT1(4). \\
\noindent \textbf{--} We achieved $\approx$ 61\% accuracy with AlexNet, with very low precision and recall for non-zero classes; hence we omit it. 

In summary, due to the sparsity of the data and class imbalance, we were unable to get good accuracy, precision, and recall values for greater than or equal to three classes. Since our partner agency is interested primarily in knowing about high conflict regions at 4x4 granularity, 2 classes are sufficient. With the hierarchical classification method, we have been able to achieve 75.7\% accuracy for 4x4 and accuracy of 80.4\% for the full hierarchical model with high precision ($\geq 0.76$) and recall ($\geq 0.66$).

\begin{table}[ht]
    \centering
    {\small \begin{tabular}[width=\linewidth]{|c||c|c|c|}
    \hline
    Model (Classes) & N2 (3) & N5 (2) & N6 (2)\\
    \hline
    \hline
    4$\times$4 Accuracy & 60.3 \% & 62.6 \% & \textbf{75.7} \% \\
    \hline
    4$\times$4 Precision & 0.22 & 0.78 & \textbf{0.85} \\
    \hline
    4$\times$4 Recall & 0.40 & 0.43 & \textbf{0.66} \\
    \hline
    \hline
    HM Accuracy & 71.2 \% & 76.9 \% & \textbf{80.4} \%\\
    \hline
    HM Precision & 0.49 & 0.74 & \textbf{0.76}\\
    \hline
    HM Recall & 0.68 & 0.69 & \textbf{0.76}\\
    \hline
    \end{tabular}}
        \caption{\small Results for the (4 km $\times$ 4 km) level and the full HM}
    \label{tab:full_acc}
    \vspace{-0.2in}
\end{table}
\begin{figure}[htbp]
    \centering
    \includegraphics[width=0.2\linewidth]{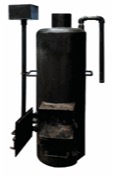}
    \includegraphics[width=0.6\linewidth]{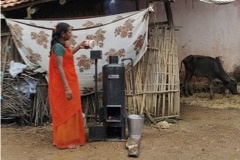}
    \caption{\small The "Bumbb" water heater}
    \label{fig:bumbb}
    \vspace{-0.2in}
\end{figure}
\section{Pilot Deployment}
\label{sec:pilot}
Our work is motivated by real-world problems hence safely testing and deploying these technologies is of critical importance. We are in the process of deploying our AI models on the field and gradually scaling our operations. Clusters of villages that surround zones/grids with high predicted conflicts (according to our methods) were identified and shortlisted. The shortlisted villages will receive interventions to reduce human wildlife conflicts. The first intervention is the provision of a water heater (Figure 4) to eliminate the need to go into the forest for fetching firewood (an extremely high-frequency activity with a high probability of conflict with wildlife). The sequential rollout across clusters (randomly chosen) lends itself as a natural experiment to be monitored. As a control-treatment set up another cluster that is yet to receive water heater will also be monitored. This will provide an on-field assessment to monitor the efficacy of the intervention. 

\bibliography{acmart}

\begin{thebibliography}{34}
\providecommand{\natexlab}[1]{#1}

\bibitem[{Sat()}]{Satellit59:online}
 ????
\newblock ESRI Satellite Maps.
\newblock \url{https://www.esri.com/en-us/maps-we-love/gallery/satellite-map}.
\newblock (Accessed on 01/31/2021).

\bibitem[{Ove()}]{Overview61:online}
 ????
\newblock Google Maps Static API.
\newblock
  \url{https://developers.google.com/maps/documentation/maps-static/overview}.
\newblock (Accessed on 01/31/2021).

\bibitem[{Allgower and Georg(2012)}]{allgower2012numerical}
Allgower, E.~L.; and Georg, K. 2012.
\newblock \emph{Numerical continuation methods: an introduction}, volume~13.
\newblock Springer Science \& Business Media.

\bibitem[{Behari et~al.(2021)Behari, Bondi, Golden, Randriamady, and
  Tambe}]{1562172}
Behari, N.; Bondi, E.; Golden, C.~D.; Randriamady, H.~J.; and Tambe, M. 2021.
\newblock Satellite-Based Food Market Detection for Micronutrient Deficiency
  Prediction.
\newblock \emph{W3PHIAI-21 at AAAI}.

\bibitem[{Bengio et~al.(2009)Bengio, Louradour, Collobert, and
  Weston}]{bengio2009curriculum}
Bengio, Y.; Louradour, J.; Collobert, R.; and Weston, J. 2009.
\newblock Curriculum learning.
\newblock In \emph{ICML}, 41--48.

\bibitem[{Bharucha(2002)}]{bharucha2002biodiversity}
Bharucha, E. 2002.
\newblock \emph{The Biodiversity of India}, volume~1.
\newblock Mapin Publishing Pvt Ltd.

\bibitem[{Bondi et~al.()Bondi, Perrault, Fang, Rice, Golden, and
  Tambe}]{bondimapping}
Bondi, E.; Perrault, A.; Fang, F.; Rice, B.~L.; Golden, C.~D.; and Tambe, M.
  ????
\newblock Mapping for Public Health: Initial Plan for Using Satellite Imagery
  for Micronutrient Deficiency Prediction.

\bibitem[{Buchholtz et~al.(2020)Buchholtz, Stronza, Songhurst, McCulloch, and
  Fitzgerald}]{buchholtz2020using}
Buchholtz, E.~K.; Stronza, A.; Songhurst, A.; McCulloch, G.; and Fitzgerald,
  L.~A. 2020.
\newblock Using landscape connectivity to predict human-wildlife conflict.
\newblock \emph{Biological Conservation}, 248: 108677.

\bibitem[{Fang et~al.(2016)Fang, Nguyen, Pickles, Lam, Clements, An, Singh,
  Tambe, Lemieux et~al.}]{fang2016deploying}
Fang, F.; Nguyen, T.~H.; Pickles, R.; Lam, W.~Y.; Clements, G.~R.; An, B.;
  Singh, A.; Tambe, M.; Lemieux, A.; et~al. 2016.
\newblock Deploying PAWS: Field Optimization of the Protection Assistant for
  Wildlife Security.
\newblock In \emph{AAAI}, volume~16, 3966--3973.

\bibitem[{Forest Survey~of India(2019)}]{fsi2019}
Forest Survey~of India, {\ }. M. o. E. {\ }. F. . C.~C. 2019.
\newblock \emph{India State of Forest Report 2019}, volume~1.
\newblock Forest Survey of India.

\bibitem[{Glorot and Bengio(2010)}]{glorot2010understanding}
Glorot, X.; and Bengio, Y. 2010.
\newblock Understanding the difficulty of training deep feedforward neural
  networks.
\newblock In \emph{AISTATS}, 249--256.

\bibitem[{Gorelick et~al.(2017)Gorelick, Hancher, Dixon, Ilyushchenko, Thau,
  and Moore}]{gorelick2017google}
Gorelick, N.; Hancher, M.; Dixon, M.; Ilyushchenko, S.; Thau, D.; and Moore, R.
  2017.
\newblock Google Earth Engine: Planetary-scale geospatial analysis for
  everyone.
\newblock \emph{Remote sensing of Environment}, 202: 18--27.

\bibitem[{Goswami et~al.(2015)Goswami, Medhi, Nichols, and
  Oli}]{goswami2015mechanistic}
Goswami, V.~R.; Medhi, K.; Nichols, J.~D.; and Oli, M.~K. 2015.
\newblock Mechanistic understanding of human--wildlife conflict through a novel
  application of dynamic occupancy models.
\newblock \emph{Conservation Biology}, 29(4): 1100--1110.

\bibitem[{Grover and Leskovec(2016)}]{grover2016node2vec}
Grover, A.; and Leskovec, J. 2016.
\newblock node2vec: Scalable feature learning for networks.
\newblock In \emph{KDD}, 855--864.

\bibitem[{Guo et~al.(2020)Guo, Xu, Cronin, Okeke, Plumptre, and
  Tambe}]{guo2020enhancing}
Guo, R.; Xu, L.; Cronin, D.; Okeke, F.; Plumptre, A.; and Tambe, M. 2020.
\newblock Enhancing Poaching Predictions for Under-Resourced Wildlife
  Conservation Parks Using Remote Sensing Imagery.
\newblock \emph{arXiv preprint arXiv:2011.10666}.

\bibitem[{Hoerl and Kennard(1970)}]{hoerl1970ridge}
Hoerl, A.~E.; and Kennard, R.~W. 1970.
\newblock Ridge regression: Biased estimation for nonorthogonal problems.
\newblock \emph{Technometrics}, 12: 55--67.

\bibitem[{Hosmer~Jr, Lemeshow, and Sturdivant(2013)}]{hosmer2013applied}
Hosmer~Jr, D.~W.; Lemeshow, S.; and Sturdivant, R.~X. 2013.
\newblock \emph{Applied logistic regression}, volume 398.
\newblock John Wiley \& Sons.

\bibitem[{Jean et~al.(2016)Jean, Burke, Xie, Davis, Lobell, and
  Ermon}]{jean2016combining}
Jean, N.; Burke, M.; Xie, M.; Davis, W.~M.; Lobell, D.~B.; and Ermon, S. 2016.
\newblock Combining satellite imagery and machine learning to predict poverty.
\newblock \emph{Science}, 353(6301): 790--794.

\bibitem[{Kenney and Keeping(1962)}]{kenney1962linear}
Kenney, J.~F.; and Keeping, E. 1962.
\newblock Linear regression and correlation.
\newblock \emph{Mathematics of statistics}, 1: 252--285.

\bibitem[{Kingma and Ba(2014)}]{kingma2014adam}
Kingma, D.~P.; and Ba, J. 2014.
\newblock Adam: A method for stochastic optimization.
\newblock \emph{arXiv preprint arXiv:1412.6980}.

\bibitem[{Krizhevsky, Sutskever, and Hinton(2017)}]{krizhevsky2017imagenet}
Krizhevsky, A.; Sutskever, I.; and Hinton, G.~E. 2017.
\newblock Imagenet classification with deep convolutional neural networks.
\newblock \emph{Communications of the ACM}, 60(6): 84--90.

\bibitem[{Kumar and Mutanga(2018)}]{kumar2018google}
Kumar, L.; and Mutanga, O. 2018.
\newblock Google Earth Engine applications since inception: Usage, trends, and
  potential.
\newblock \emph{Remote Sensing}, 10(10): 1509.

\bibitem[{LeCun, Bengio, and Hinton(2015)}]{lecun2015deep}
LeCun, Y.; Bengio, Y.; and Hinton, G. 2015.
\newblock Deep learning.
\newblock \emph{nature}, 521(7553): 436--444.

\bibitem[{Lute et~al.(2016)Lute, Navarrete, Nelson, and Gore}]{lute2016moral}
Lute, M.~L.; Navarrete, C.~D.; Nelson, M.~P.; and Gore, M.~L. 2016.
\newblock Moral dimensions of human--wildlife conflict.
\newblock \emph{Conservation Biology}, 30(6): 1200--1211.

\bibitem[{McCleery(2009)}]{mccleery2009improving}
McCleery, R.~A. 2009.
\newblock Improving attitudinal frameworks to predict behaviors in
  human--wildlife conflicts.
\newblock \emph{Society and natural Resources}, 22(4): 353--368.

\bibitem[{Naha et~al.(2019)Naha, Sathyakumar, Dash, Chettri, and
  Rawat}]{naha2019}
Naha, D.; Sathyakumar, S.; Dash, S.; Chettri, A.; and Rawat, G.~S. 2019.
\newblock Assessment and prediction of spatial patterns of human-elephant
  conflicts in changing land cover scenarios of a human-dominated landscape in
  North Bengal.
\newblock \emph{PLOS ONE}, 14(2): 1--19.

\bibitem[{Pandey, Agarwal, and Krishnan(2018)}]{pandey2018multi}
Pandey, S.; Agarwal, T.; and Krishnan, N.~C. 2018.
\newblock Multi-task deep learning for predicting poverty from satellite
  images.
\newblock In \emph{AAAI}, volume~32.

\bibitem[{Pinjarkar(2019)}]{pinjakartimes}
Pinjarkar, V. 2019.
\newblock Man animal conflict stalks candidates in forest areas.
\newblock \emph{TOI}.

\bibitem[{Sharma et~al.(2020)Sharma, Chettri, Uddin, Wangchuk, Joshi, Tandin,
  Pandey, Gaira, Basnet, Wangdi et~al.}]{sharma2020mapping}
Sharma, P.; Chettri, N.; Uddin, K.; Wangchuk, K.; Joshi, R.; Tandin, T.;
  Pandey, A.; Gaira, K.~S.; Basnet, K.; Wangdi, S.; et~al. 2020.
\newblock Mapping human--wildlife conflict hotspots in a transboundary
  landscape, Eastern Himalaya.
\newblock \emph{Global Ecology and Conservation}, 24: e01284.

\bibitem[{Suykens and Vandewalle(1999)}]{suykens1999least}
Suykens, J.~A.; and Vandewalle, J. 1999.
\newblock Least squares support vector machine classifiers.
\newblock \emph{Neural processing letters}, 9(3): 293--300.

\bibitem[{Treves and Santiago-{\'A}vila(2020)}]{treves2020myths}
Treves, A.; and Santiago-{\'A}vila, F.~J. 2020.
\newblock Myths and assumptions about human-wildlife conflict and coexistence.
\newblock \emph{Conservation Biology}, 34(4): 811--818.

\bibitem[{Woodroffe, Thirgood, and Rabinowitz(2005)}]{woodroffe2005people}
Woodroffe, R.; Thirgood, S.; and Rabinowitz, A. 2005.
\newblock \emph{People and wildlife, conflict or co-existence?}
\newblock 9. Cambridge University Press.

\bibitem[{Xu et~al.(2020)Xu, Gholami, Mc~Carthy, Dilkina, Plumptre, Tambe,
  Singh, Nsubuga, Mabonga, Driciru et~al.}]{xu2020stay}
Xu, L.; Gholami, S.; Mc~Carthy, S.; Dilkina, B.; Plumptre, A.; Tambe, M.;
  Singh, R.; Nsubuga, M.; Mabonga, J.; Driciru, M.; et~al. 2020.
\newblock Stay Ahead of Poachers: Illegal Wildlife Poaching Prediction and
  Patrol Planning Under Uncertainty with Field Test Evaluations (Short
  Version).
\newblock In \emph{ICDE}, 1898--1901. IEEE.

\bibitem[{Zhang(2004)}]{zhang2004solving}
Zhang, T. 2004.
\newblock Solving large scale linear prediction problems using stochastic
  gradient descent algorithms.
\newblock In \emph{ICML}, 116.

\end{thebibliography}

\end{document}